\documentclass[11pt,a4paper]{article}
\usepackage[hyperref]{acl2019}
\usepackage{times}
\usepackage{latexsym}
\usepackage{microtype}
\usepackage{url}
\usepackage{amsmath}
\usepackage{amsfonts}
\usepackage{bbm}
\usepackage{booktabs}
\usepackage{tikz}
\usepackage{adjustbox}
\usepackage{amsbsy}
\usetikzlibrary{bayesnet}
 \usepackage{booktabs}
 \usepackage{longtable}
 \usepackage{array}
 \usepackage{multirow}
 \usepackage{wrapfig}
 \usepackage{float}
 \usepackage{colortbl}
 \usepackage{pdflscape}
 \usepackage{tabu}
 \usepackage{threeparttable}
 \usepackage{threeparttablex}
 \usepackage[normalem]{ulem}
  \usepackage{makecell}
  \usepackage{xcolor}

\newcommand*{\mybox}[1]{%
  \fcolorbox{white}{white}{\raisebox{0pt}[0.5\baselineskip][0.05\baselineskip]{%
      \hbox to 1.8cm{\hss#1\hss}}}}

\newcommand*{\myboxgrey}[1]{%
  \fcolorbox{gray!25}{gray!25}{\raisebox{0pt}[0.5\baselineskip][0.05\baselineskip]{%
    \hbox to 1.8cm{\hss#1\hss}}}}

\definecolor{awesome}{rgb}{1.0, 0.13, 0.32}
\definecolor{blue(pigment)}{rgb}{0.2, 0.2, 0.6}
\definecolor{darkcyan}{rgb}{0.0, 0.55, 0.55}
\definecolor{electricviolet}{rgb}{0.56, 0.0, 1.0}

\DeclareFontFamily{U}{rcjhbltx}{}
\DeclareFontShape{U}{rcjhbltx}{m}{n}{<->rcjhbltx}{}
\DeclareSymbolFont{hebrewletters}{U}{rcjhbltx}{m}{n}

\usepackage{cleveref}
\crefname{section}{\S}{\S\S}
\Crefname{section}{\S}{\S\S}
\crefname{table}{Table}{}
\crefname{figure}{Figure}{}
\crefname{algorithm}{Algorithm}{}
\crefname{equation}{eq.}{}
\crefname{appendix}{App.}{}
\crefname{ExNo}{Sentence}{}
\crefformat{section}{\S#2#1#3}  %

\usepackage{todonotes}
\makeatletter
\newcommand*\iftodonotes{\if@todonotes@disabled\expandafter\@secondoftwo\else\expandafter\@firstoftwo\fi}  %
\makeatother

\newcommand{\word}[1]{\textit{#1}}
\newcommand{\lex}[1]{\textsc{#1}}

\newcommand{\al}{\mathbf{a}}

\newcommand{\defn}[1]{\textbf{#1}}
\newcommand{\fm}[1]{\textsl{#1}}

\newcommand{\ptheta}{p_{{\boldsymbol \theta}}}

\newcommand{\calA}{{\cal A}}

\newcommand{\calC}{{\cal C}}
\newcommand{\bll}{\ensuremath{\boldsymbol{\ell}}}
\newcommand{\Pb}{\ensuremath{\mathbb{P}}}

\aclfinalcopy %
\def\aclpaperid{1374} %

\title{Morphological Irregularity Correlates with Frequency}

\author{Shijie Wu \\
  Department of Computer Science \\
  Johns Hopkins University \\
  Baltimore, US \\
  \texttt{shijie.wu@jhu.edu} \\\And
  Ryan Cotterell \\
  The Computer Laboratory \\
  University of Cambridge \\
  Cambridge, UK \\
  \texttt{rdc42@cam.ac.uk}
  \\\And
  Timothy J. O'Donnell \\
  Department of Linguistics, Mila  \\
  McGill University \\
  Montr\'{e}al, Canada \\
  \texttt{timothy.odonnell@mcgill.ca}\\}

\date{}

\begin{document}
\maketitle
\begin{abstract}
We present a study of morphological irregularity. Following recent
  work, we define an information-theoretic measure of irregularity
  based on the predictability of forms in a language. Using a neural
  transduction model, we estimate this quantity for the forms in 28
  languages. We first present several validatory and exploratory
  analyses of irregularity. We then show that our analyses provide
  evidence for a correlation between irregularity and frequency:
  higher frequency items are more likely to be irregular and irregular
  items are more likely be highly frequent. To our knowledge, this
  result is the first of its breadth and confirms longstanding proposals from the linguistics
  literature. The correlation is more robust when aggregated at the
  level of whole paradigms---providing support for models of
  linguistic structure in which inflected forms are unified by
  abstract underlying stems or lexemes.
  Code is available at \url{https://github.com/shijie-wu/neural-transducer}.

\end{abstract}

\section{Introduction}
Irregularity is a pervasive phenomenon in the inflectional morphology
of the world's languages and raises a number of questions about
language design, learnability, and change. Nevertheless, irregularity
remains an understudied phenomenon and many basic questions remain
unanswered \citep[][]{kiefer.f:2000,stolz.t:2012a}. Do all languages
exhibit irregularity? What is the relationship between irregularity
and frequency? Is irregularity best thought of as a property of
individual forms, or a property of more abstract objects like
morphological paradigms?  In this paper, we examine these questions,
focusing in particular on the relationship between irregularity and
frequency.

One of the fundamental challenges in studying irregularity is defining
the phenomenon in a way that is applicable across languages. We begin the
paper by addressing this question. First, we formalize the problem of
inflectional morphology and present a novel, information-theoretic
measure of the degree of irregularity of an inflected form. This
definition builds on recent work that defines (ir)regularity in terms
of the probabilistic predictability of a form given the rest of the
language \citep[][]{cotterell.r:2018,ackerman.f:2013}.  Making use of
a state-of-the-art model of morphological inflection, we estimate our
measure of irregularity across a large number of word forms from 28
languages drawn from the UniMorph database
\citep[][]{kirov2018unimorph}. Based on these estimates we perform
three studies. First, we validate our estimates by examining the
predictions on English past tense forms---showing that the model's
predicts accord with human judgements of irregularity. We also examine
the overall rate of accuracy of our model. Second, we examine the
degree of irregularity across languages, showing that the model
predicts wide variance in the average amount of irregularity between
the languages in our sample. Finally, we provide empirical evidence
for a correlation between irregularity and frequency across
languages. While this relationship has been observed for individual languages \citep[e.g., English:][]{marcus.g:1992,bybee.j:1985}, this is the first confirmation of the effect across this many languages.
This result is especially relevant given recent discussions calling the relationship into question  \citep[e.g.,][]{fratini.v:2014,yang.c:2016}. We find,
furthermore, that the correlation between irregularity and frequency
is much more robust when irregularity is considered as a property of
whole \fm{lexemes} (or \fm{stems}/\fm{paradigms}) rather than as a
property of individual word forms. We discuss the implications of
these findings.

\section{Formalizing Inflectional Morphology}\label{sec:formalization}

In this work, each word type is represented as a triple consisting of
the following components:

\begin{itemize}
  \setlength\itemsep{0.5em}

\item A \defn{lexeme}\footnote{This terminology is characteristic of
    \fm{word-and-paradigm} approaches to morphology. In
    \fm{item-and-arrangement} approaches, this might be called the
    \fm{stem} \citep[][]{hockett.c:1954}. } $\ell$: An arbitrary
  integer or string that indexes an abstract word (e.g., \lex{go},
  which provides an index to forms of the verb \word{go} such as
  \word{goes} and \word{went}).

\item A \defn{slot} $\sigma$: An arbitrary integer, string, or more
  structured object that indicates how the word is inflected (e.g.,
  $[\textit{pos}$$=$$\text{v},\textit{tns}$$=$$\text{past},\textit{person}$$=$$\text{3rd},\textit{num}$$=$$\text{sg}]$
  for the form \word{went}).

\item A \defn{surface form} $w$: A string over a fixed phonological or
  orthographic alphabet $\Sigma$ (e.g., \word{went}).
  
\end{itemize}

A \defn{paradigm} $\bll$ (boldface $\ell$) is a lexeme-specific map
from slots to surface forms for lexeme $\ell$.\footnote{See
  \citep[][Part II]{baerman.m:2015} for a tour of alternative views of
  inflectional paradigms.}  Typically, slots are indexed by structured
entities---known as \fm{morpho-syntactic feature vectors} or
\fm{morpho-syntactic tags}---represented by a set of key-value pairs:
$\sigma
=[k_1$$=$$v_1,\ldots,k_n$$=$$v_n]$. For example, the English verb form
\word{runs}, which has the feature vector
$[\textit{tns}$$=$$\text{pres},\textit{per}$$=$$\text{3rd},
\textit{num}$$=$$\text{sing}]$.  In what follows, the keys $k_i$ and
the corresponding values $v_i$ are taken from the universal inventory,
defined by the UniMorph annotation scheme and denoted $\mathcal{M}$
\citep[][]{kirov2018unimorph}.  We use dot notation to refer to
specific forms or sets of forms in a paradigm indexed by some slot
$\lex{go}.past=$ \word{went}.
 
Given the pieces just sketched, a complete model of inflectional
morphology will specify a joint distribution over surface forms,
lexemes, and slots, that is $\mathbb{P}(w, \ell, \sigma)$, or one of
its associated conditional distributions, such as
$\mathbb{P}(\ell, \sigma \mid w)$---the distribution over lexemes and
features, given a surface form; or
$\mathbb{P}(w\mid \ell, \sigma)$---the conditional probability of a
surface form given a lexeme and inflectional features. In this paper,
we will focus on the latter, defining a probabilistic model to
approximate this distribution and using that to estimate degrees of
irregularity.

\section{Operationalizing Irregularity}\label{sec:irregularity}
The informal distinction between regular and irregular forms is an
important one for many theories of grammar
\citep[e.g.,][]{siegel.d:1974}, language processing
\citep[e.g.,][]{hay.j:2003}, and language acquisition
\citep[e.g.,][]{pinker.s:1999,marcus.g:1992,mcclelland.j:2002,mcclelland.j:2002a,pinker.s:2002,pinker.s:2002a,rumelhart.d:1986,prasada.s:1993,pinker.s:1988}. However,
there have been few proposals for how the notion can be characterized
precisely or measured quantitatively.

Clearly, the regularity of a form (or rule) can only be defined with
respect to the language as a whole---what makes something irregular is
that it does not behave in the way that would be expected given other
forms in the language. But what is meant by \emph{expected}? Here, we
follow recent work by defining the notion of expectedness in terms of
a probabilistic model of inflection which approximates
$\mathbb{P}(w\mid \ell, \sigma)$
\citep[][]{cotterell.r:2018,ackerman.f:2013}. However, there remains a
wrinkle.  A form like \word{went} is highly expected as the past tense
of \lex{go} for an adult speaker of English, but is also
irregular. How do we capture this?

We take the correct notion of expectedness to be the expectedness of
the word form treated \emph{as if} it were the first instance of that
lexeme which had been observed.  Thus, we base our measures of
regularity on the conditional probability of a word type $w$ given the
rest of the forms in the language with the target lexeme removed.

\begin{equation}
\Pb(w \mid \ell, \sigma, \mathcal{L}_{-\ell})
\end{equation}

Of course, since the target language $\mathcal{L}$ is generally
infinite, we will need to make use of some model-based estimate of
this probability $\ptheta(w \mid \ell, \sigma, \mathcal{L}_{-\ell})$.
In essence, our definition of irregularity is based on
\fm{wug-testing} \citep[][]{berko.j:1958} such a probabilistic model
to see how robustly it generalizes to the target form $w$. In
practice, we will estimate this quantity by performing a holdout
evaluation of the target form under our model.

More irregular forms will tend to have a lower \word{wug}-test
probability $\Pb(w \mid \ell, \sigma, \mathcal{L}_{-\ell})$ than most
regular forms. However, the absolute value of such a probability is
not directly interpretable.  To turn these probabilities into
interpretable values which directly measure irregularity, we take the
negative log odds of the probability of the correct word form.

\begin{equation}\label{eq:log-odds}
\iota(w) = -\log \left[ \frac{\Pb(w \mid \ell, \sigma, \mathcal{L}_{-\ell})}{1-\Pb(w \mid
  \ell, \sigma, \mathcal{L}_{-\ell})} \right]
\end{equation}

We refer to this quantity as the \fm{degree of irregularity} of a
form. If probability of the correct form $w$ is exactly $0.5$, then
\cref{eq:log-odds} will be $0$. However, if
$\Pb(w \mid \ell, \sigma, \mathcal{L}_{-\ell}) > \sum_{w^\prime \neq
  w} \Pb(w^\prime \mid \ell, \sigma, \mathcal{L}_{-\ell})$, then
\cref{eq:log-odds} will be negative. Otherwise, the quantity is
positive. In other words, the metric is more strongly positive when a
form is less predictable given other forms in the language and more
strongly negative when a form is more strongly predictable.  The
midpoint at $0$ occurs when there is an equal amount of probability
mass on the correct form and all other forms.

Note that this definition of $\iota$ neatly addresses several
challenges in studying the notion of (ir)regularity. First, it doesn't
require us to define a binary notion of regular versus irregular or
even to explicitly define any such notion at all---a model may treat
regularity as an implicit rather than explicit feature of a form or
paradigm. Second, and relatedly, we do not require data annotated with
the regularity of forms to train or test our model. Third, this
definition inherently captures the idea of degree of regularity, for
instance, capturing the distinction between wholly suppletive forms
such as \word{went} and semi-productive inflectional classes such as
\word{ring}/\word{rang}, \word{sing}/\word{sang}, etc. Fourth and
finally, regularity is known to be correlated with other features of
morphological structure, such as productivity. Our definition
sidesteps the tricky issue of disentangling these different properties
of inflection.

Note that our definition of $\iota$ conditions on
$\mathcal{L}_{-\ell}$---the language without the target
\emph{lexeme}---rather than on $\mathcal{L}_{-w}$---the language
without the target \emph{word}. Thus, we are measuring the probability
that the model will generalize to the correct form without any
evidence of a lexeme at all. Thus, we rule out predictability that
comes from similar forms within a paradigm $\bll$. For example, in our
approach a model cannot make use of the irregularity of the past tense
form \word{ring} to guess that the past participle form was more
likely to be \word{rung}. We discuss the implications of this
assumption in more detail below \cref{sec:frequency}.

\section{Modeling Morphological Inflection}
Our goal is to estimate
$\Pb(w \mid \ell, \sigma, \mathcal{L}_{-\ell})$ from data. We do this
by using a structured probabilistic model of string transduction which
we call $\ptheta$. In the following sections, we describe this model,
how we handle syncretism in the model, our training (holdout and test)
scheme, and our estimates of the degree of irregularity $\iota$.

\subsection{A Lemma-Based Model}

\begin{figure}
  \centering
        \begin{tikzpicture}[scale=0.75, every node/.style={transform shape}]
          \node[obs] (lemma) {$\text{\myboxgrey{poner}}$} ; %
          \node[latent, left=of lemma, xshift=-0.0cm] (m1) {$\text{\mybox{pongo}}$} ; %
          \node[latent, below left=of lemma, yshift=-0.0cm] (m2) {$\text{\mybox{pongas}}$} ; %
          \node[latent, below=of lemma, yshift=0.0cm] (m3) {$\text{\mybox{ponga}}$} ; %
          \node[latent, below right=of lemma, yshift=-0.0cm] (m4) {$\text{\mybox{pongan}}$} ; %
          \node[latent, right=of lemma, yshift=-0.0cm] (m5) {$\text{\mybox{pondr{\'i}as}}$} ; %
          \node[latent, above right=of lemma, yshift=-0.0cm] (m6) {$\text{\mybox{pondr{\'i}ais}}$} ; %
          \node[latent, above left=of lemma, yshift=-0.0cm] (m7) {$\text{\mybox{pondr{\'ian}}}$} ; %
          \node[latent, above=of lemma, yshift=-0.0cm] (m8) {$\text{\mybox{pondr{\'i}as}}$} ; %
          \edge {lemma} {m1, m2, m3, m4, m5, m6, m7, m8};
        \end{tikzpicture}
        \caption{Lemma paradigm tree}
        \label{fig:lemma}
\end{figure}

In linguistic morphology, a major division is between
\fm{item-and-arrangement} or \fm{morpheme-based} models and
\fm{word-and-paradigm} or \fm{word-based} models
\citep[][]{hockett.c:1954}.  Following \citep[][]{cotterell.r:2017},
we adopt a word-based approach. To do this, we designate a unique
surface form for each paradigm $\bll$ known as the \fm{lemma}. The
lemma is associated with a slot which we notate $\check{\sigma}$:
$\bll.\check{\sigma} \in \Sigma^*$. The lemma can be thought of as a
dictionary or citation form of a word and is traditionally chosen by
lexicographers of a language. For example, in many Western European
languages the lemma of verb forms is the
infinitive. \Cref{fig:lemma} shows several of the forms of the
Spanish verb \word{poner} (``to put'') organized around the lemma
form. In what follows, we use the lemma to identify 
lexemes, and wherever a probability distribution would condition on the
abstract lexeme $\ell$ we instead condition on the lemma
$\bll.\check{\sigma}$.

Our probabilistic model of string transduction $\ptheta$ is a
monotonic model with hard attention described in \citet[][]{wu2019exact} and
can be viewed as a graphical model over strings like the one shown in
\cref{fig:lemma}. It is expressed as follows.

\begin{equation}
  \ptheta(w \mid \bll.\check{\sigma}, \sigma, \mathcal{L}_{-\ell}) =
  \hspace{-.5cm}\sum_{\al \in \calA(w, \bll.\check{\sigma})}
  \hspace{-.3cm}\ptheta(w, \al \mid \bll.\check{\sigma}, \sigma, \mathcal{L}_{-\ell}).
\end{equation}

The definition of the model includes a sum over all monotonic
(non-crossing) alignments $\calA(w, \bll.\check{\sigma})$ between the
lemma $\bll.\check{\sigma}$ and the output surface form $w$. The inner
term of this sum is estimated using a sequence to sequence model. The
sum itself is computable in polynomial time using a variant of the
forward algorithm \citep[][]{rabiner1989tutorial}.
The model achieves state-of-the-art performance on the SIGMORPHON 2017
shared task on morphological reinflection \citep[][]{K17-2001}. We follow the hyperparameter used by \citet[][]{wu2019exact}.

\subsection{Handling Syncretism}
\label{sec:syncretism}

Many inflectional systems display \fm{syncretism}---the morphological
phenomenon whereby two slots with distinct morpho-syntactic tags may
have an identical surface form.  In contrast to many models of
inflectional morphology, we collapse syncretic forms of a word into
a single paradigm slot, thereby assuming that every every surface form
$w$ in a paradigm is distinct. An example of such a collapsed paradigm
in German is given in \cref{tab:paradigm}. Our formalization includes
a slot that merges the genitive, accusative and dative singular into a
single slot due to the word \word{Herr}.

\begin{table}
\centering
\begin{tabular}{lllll}
  \toprule
  & {\sc sg} & {\sc pl} & {\sc sg} & {\sc pl}  \\[.2em] \midrule
      {\sc nom} & \textbf{\color{electricviolet}{\word{Wort}}} & \textbf{\color{darkcyan}{\word{W{\"o}rter}}} & \word{Herr} & \textbf{\color{awesome}{\word{Herren}}} \\[.2em]
    {\sc gen} & \word{Wortes} & \textbf{\color{darkcyan}{\word{W{\"o}rter}}} & \textbf{\color{blue(pigment)}{\word{Herrn}}} & \textbf{\color{awesome}{\word{Herren}}} \\[.2em]
{\sc acc} &  \textbf{\color{electricviolet}{\word{Wort}}} & \textbf{\color{darkcyan}{\word{W{\"o}rter}}}  & \textbf{\color{blue(pigment)}{\word{Herrn}}} & \textbf{\color{awesome}{\word{Herren}}} \\[.2em]
{\sc dat}  & \word{Worte} & \word{W{\"o}rtern} & \textbf{\color{blue(pigment)}{\word{Herrn}}} & \textbf{\color{awesome}{\word{Herren}}} \\[.2em]
\bottomrule
\end{tabular}
\caption{Full paradigms for the German nouns \word{Wort} (``word'')
  and \word{Herr} (``mister'') with abbreviated and tabularized
  UniMorph annotation. The syncretic forms are bolded and colored by
  ambiguity class. Note that, while in the plural, the nominative and
  accusative are always syncretic across all paradigms, the same is
  not true in the singular.  }
\label{tab:paradigm}
\end{table}

To accomplish this we assume that each lexeme is associated with a set
of \fm{syncretism classes} denoted by $\calC^\ell$.
$\calC^\ell: \mathcal{M} \rightarrow \mathcal{M}$ is a map from a slot
$\sigma$ to a \fm{citation form slot} $\sigma^\prime$ which indexed
the canonical suface citation form for that combination of
features. $\calC^\ell$ is used to collapse paradigm cells with
identical surface forms. For instance, all forms of the lexeme
\lex{go} are realized as \word{went} in the English past tense,
regardless of person and number features; thus, for example,
$\calC^{\lex{go}}([\textit{tns}=\text{past},\textit{per}=\text{3rd},
\textit{num}=\text{sing}])=\calC^{\lex{go}}([\textit{tns}=\text{past},\textit{per}=\text{2nd},\textit{num}=\text{plural}])$. We
say that two lexemes $\ell$ and $\ell'$ are \fm{syncretically
  equivalent} if $\calC^\ell(\sigma) = \calC^{\ell'}(\sigma) $ for all
$\sigma$.  We assume the mappings $\calC^\ell$ are known and given in
advance in what follows.

We will use this syncretism-collapsed representation for all
simulations below. In particular, this assumption will allow us to
simply count the surface forms of each word in Wikipedia without
dealing with the tricky issue of assigning individual words to the
correct combination of morphosyntactic features \citep[see,][for
detailed discussion]{N18-2087}.

\subsection{Handling Derived Forms}\label{sec:snag}

As discussed above, we hold out whole lexemes, including all of their
inflected forms during training. However, derivational morphology
presents a potential challenge for this approach.  Consider the
irregular verb \word{do}/\word{did}/\word{done}. This verb appears in
a number of derived prefixed forms such as \word{redo} and
\word{undo}. These forms all inflect identically to the base form
\word{do}---for example, \word{redo}/\word{redid}/\word{redone}.\footnote{An anonymous reviewer points out that in some languages, such as Dutch, forms derived from irregular verbs become regular (e.g., \word{zeggen}/\word{zei} but \word{toezeggen}/\word{toezegde}). In those languages, it should be unnecessary to apply our heuristic approach.}  If
we train our probability model on such derived forms, it is likely to
estimate too high a \word{wug}-test probability for all forms which
are built from the shared stem.

To obviate this problem, we remove all derived forms from the data we
consider. To do so we develop a heuristic approach to isolate all
words that may have been derived from another. Note that a key
desideratum of heuristic is that it should be high precision with
respect to finding derivational transformation---we would rather
overexclude forms as potentially derivative of another, rather than
leave a derived form in the data.

We consider a lexeme $\ell^\prime$ to be derived from a lexeme $\ell$
if and only if there is a string $s \in \Sigma^+$ such that
$(\forall \sigma)[\bll^\prime.\sigma = \bll.\sigma \cdot s]$ or
$(\forall \sigma)[\bll^\prime.\sigma = s \cdot \bll.\sigma]$ where
$s \cdot t$ denotes string concatenation of strings $s$ and $t$. For
example, \lex{do} and \lex{redo} satisfy this condition, while
\lex{sing} and \lex{ring} do not. We perform a search for candidate
$s$ for all pairs of lexemes in each language and remove all
$\ell^\prime$ that meet this criterion.

\subsection{Measuring Irregularity}
With the above definitions in place, we can define an approximation to
our degree of irregularity $\iota$.

\begin{equation}\label{eq:log-odds-specific}
  \iota(w) =  -\log \frac{\ptheta(w \mid \bll.\check{\sigma}, \sigma, \mathcal{L}_{-\ell})}{1 -
      \ptheta(w \mid \bll.\check{\sigma}, \sigma, \mathcal{L}_{-\ell})}
  \end{equation}

  In our analyses below, we will also wish to measure the irregularity
  of lexemes as a whole. To do this, we take the average irregularity
  score over the entire paradigm $\bll$.
  
  \begin{equation}
    \label{eq:avg-irregularity}
    \iota(\ell) = \frac{\sum_{\{(w, \sigma, \ell) \in \bll\,\,\mid\,\, w \neq
        \bll.\check{\sigma} \}} -\log\frac{ \ptheta(w \mid
        \bll.\check{\sigma}, \sigma, \mathcal{L}_{-\ell})}{1 - \ptheta(w \mid
        \bll.\check{\sigma}, \sigma, \mathcal{L}_{-\ell})}}{|\bll|-1}
\end{equation}

\section{Studies of Irregularity}
The empirical portion of our work consists of three studies. We first
validate and examine the accuracy of the model
(\cref{sec:validation}). Second, we examine the distribution of
irregularity across the languages in our sample
(\cref{sec:irregularity-across}). Finally, we examine the correlation
between irregularity and frequency (\cref{sec:frequency}). Before
presenting these studies we first give an overview of the data and
simulations common to all of them.

\subsection{Simulations}\label{sec:simulations}

\paragraph{Data Provenance.}
All word forms, paradigms, and morphosyntactic features are taken from
the UniMorph project \cite{kirov2018unimorph}. Specifically, we
examine the following 28 languages: Albanian, Arabic, Armenian,
Basque, Bulgarian, Czech, Danish, Dutch, English, Estonian, French,
German, Hebrew, Hindi, Irish, Italian, Latvian, Persian, Polish,
Portuguese, Romanian, Russian, Spanish, Swedish, Turkish, Ukrainian,
Urdu, and Welsh. The languages come from 4 stocks (Indo-European,
Afro-Asiastic, Finno-Urgic and Turkic) with Basque, a language
isolate, included as well. Although this sample represents a
reasonable degree of typological diversity, the Indo-European family
is overrepresented in the UniMorph dataset, as is the case for most
current multilingual corpora. However, within the Indo-European
family, we consider a diverse set of subfamilies: Albanian, Armenian,
Slavic, Germanic, Romance, Indo-Aryan, Baltic, and Celtic. For each
subfamily, we sample subset of languages randomly.

All of our form-level frequencies were computed from
Wikipedia.\footnote{Wikipedia data retrieved on Feb 1$^\text{st}$,
  2019.}  Lexeme counts are achieved by summing over all entries in
the paradigm associated with a lexeme. In all simulations, we predict
the orthographic form of the target word $w$ from the orthographic form
of the lemma $\bll.\check{\sigma}$ as a proxy for phonological
transcriptions which do not exist for our all languages in UniMorph.

\paragraph{Lexeme-Based Cross Validation.}
In the studies that follow, we train a separate instance of our model
on the forms in each language using the following procedure. We first
remove morphologically-complex forms that are derived from other
lemmas in the corpus using the heuristic technique described in
\cref{sec:snag}. We then randomly assign the remaining lexemes of each
language to one of ten splits. Note that each split will contain all
of the forms associated with each lexeme and a lexeme will never be
divided across splits.  We then perform 10-fold cross-validation,
training the model $\ptheta$ on 8 splits, tuning on one of the
remaining two splits, and testing on the final remaining split. Note
that this approach to cross-validation allows us to approximate
$\mathcal{L}_{-\ell}$ without the costly procedure of retraining for
every held-out lexeme.  However, also note that this approach has a
potential confound. Lexemes can often be grouped into \fm{inflectional
  classes} in which all lexemes mark different slots in the same
way. For example, verbs such as \word{sing}/\word{sang}/\word{sung}
and \word{ring}/\word{rang}/\word{rung} form an inflectional class in
English. Inflectional classes vary in their size and regularity
\citep[][]{stump.g:2001}. If all or most lexemes in the same irregular
inflectional class end up together in the test split under our
approach, we may systematically overestimate their irregularity.

\subsection{Validation and Accuracy}
\subsubsection{Validation on English Verbs}\label{sec:validation}
\begin{table}
    \centering
    \begin{tabular}{cc} \toprule \textbf{
        \citet[][]{albright2003rules}} & \textbf{\citet[][]{odonnell.t:2015}}\\
      \midrule 0.670 &0.559 \\
      \bottomrule
    \end{tabular}
    \caption{Validation of our irregularity metric. Spearman's $\rho$
      between gold-standard irregularity annotations from
      \citet[][]{albright2003rules} and \citet[][]{odonnell.t:2015} and our irregularity
      metric.}
    \label{tab:study1}
  \end{table}
  
  The first question we wish to ask is whether the irregularity
  predictions made by our model are consistent with human
  intuitions. To answer this question, we examine the predictions of
  our model on the English past tense---a morphological system which
  has been intensely studied for decades \citep[see][for
  overview]{pinker.s:1999} and for which there is general agreement
  about which forms are regular or irregular. We make use of the
  databases of \citet[][]{albright2003rules} which consists of 4039
  English verb forms and the dataset of \citet[][]{odonnell.t:2015}
  which consists of 15202 verb forms, both hand-annotated for
  irregularity by experts.

  We present our results in \cref{tab:study1}. We find that our
  measure of irregularity strongly correlates with human intuitions on
  English verbs.  We take this as tentative validation of our
  metric. Future work will investigate the linguistic plausibility of
  our metric on a greater diversity of languages.

\subsubsection{\word{Wug}-Test Accuracy}\label{sec:accuracy}

\begin{table}[H]
\centering
\resizebox{\linewidth}{!}{
\begin{tabular}{llrrrr}
\toprule
Language & Family & Avg. Accuracy & Lexemes & Forms & Avg. Forms/Lexeme\\
\midrule
\rowcolor{gray!6}  Albanian & Indo-European & 0.83 & 537 & 26993 & 50.4\\
Arabic & Semitic & 0.63 & 3559 & 89879 & 25.5\\
\rowcolor{gray!6}  Armenian & Indo-European & 0.95 & 4614 & 144841 & 31.4\\
\textbf{Basque} & \textbf{Isolate} & \textbf{0.01} & \textbf{26} & \textbf{10382} & \textbf{441.9}\\
\rowcolor{gray!6}  Bulgarian & Slavic & 0.94 & 2042 & 36007 & 17.7\\
\addlinespace
Czech & Slavic & 0.92 & 4470 & 61251 & 13.8\\
\rowcolor{gray!6}  Danish & Germanic & 0.65 & 2580 & 19968 & 7.8\\
Dutch & Germanic & 0.94 & 3932 & 20680 & 5.3\\
\rowcolor{gray!6}  English & Germanic & 0.95 & 9915 & 40210 & 4.1\\
Estonian & Uralic & 0.79 & 817 & 31711 & 38.9\\
\addlinespace
\rowcolor{gray!6}  French & Romance & 0.86 & 5378 & 195638 & 37.4\\
German & Germanic & 0.92 & 14739 & 69190 & 4.7\\
\rowcolor{gray!6}  Hebrew & Semitic & 0.78 & 492 & 11240 & 23.3\\
Hindi & Indo-Aryan & 0.74 & 254 & 26404 & 104.0\\
\rowcolor{gray!6}  Irish & Celtic & 0.85 & 6527 & 69551 & 10.7\\
\addlinespace
Italian & Romance & 0.99 & 6495 & 269908 & 41.9\\
\rowcolor{gray!6}  Latvian & Baltic & 0.97 & 5347 & 60146 & 11.9\\
Persian & Iranian & 0.70 & 271 & 26336 & 98.3\\
\rowcolor{gray!6}  Polish & Slavic & 0.93 & 8317 & 106914 & 13.0\\
Portuguese & Romance & 0.98 & 2621 & 138372 & 52.9\\
\addlinespace
\rowcolor{gray!6}  Romanian & Romance & 0.78 & 3409 & 51670 & 15.3\\
Russian & Slavic & 0.95 & 19991 & 243748 & 12.2\\
\rowcolor{gray!6}  Spanish & Romance & 0.97 & 3904 & 232676 & 59.9\\
Swedish & Germanic & 0.89 & 6451 & 43118 & 6.7\\
\rowcolor{gray!6}  Turkish & Turkic & 0.85 & 2697 & 150477 & 55.9\\
\addlinespace
Ukrainian & Slavic & 0.86 & 1426 & 13844 & 9.8\\
\rowcolor{gray!6} \textbf{Urdu} & \textbf{Indo-Aryan} & \textbf{0.38} & \textbf{180} & \textbf{5581} & \textbf{31.0}\\
\textbf{Welsh} & \textbf{Celtic} & \textbf{0.41} & \textbf{179} & \textbf{9083} & \textbf{50.8}\\
\bottomrule
\end{tabular}}
\caption{Accuracy per language.  \label{table:performance}}
\end{table}

Our lexeme-based cross-validation setup differs substantially from the
form-based setup typically used to evaluate models of inflectional
morphology \citep[see, e.g.,][]{K17-2001}.  In the typical evaluation
setup, individual surface word forms are heldout, rather than all of
the forms associated with entire lexemes. This means, amongst other
things, that words from irregular lexemes will often be split between
test and train, giving models an opportunity to learn partially
productive and semi-regular patterns of inflection. Our approach
however makes this impossible by strictly assigning all forms from
each lexeme to either train or test.

It is important to ask, therefore, how well does our model predict the
forms of heldout lexemes given this stricture? The results are
displayed in \cref{table:performance}. This table displays the average
accuracy for each language in our sample as well as the number of
lexemes for that language, the total number of forms, and the average
number of forms per lexeme. The majority of languages show very high
generalization accuracy to our lexeme-based \textit{wug}-tests: 21 out
of 28 have an average accuracy of 75\% or higher. Three languages
stand out in terms of their low accuracy and are highlighted in
\cref{table:performance}: Basque, Urdu, and Welsh. These languages,
Basque especially, are characterized by smaller numbers of lexemes and
larger numbers of forms per lexeme.

In the \cref{sec:frequency}, we discuss the correlation between
irregularity and frequency. The interpretation of these results relies
on the ability of our model to accurately capture regular structure in
the inflectional systems of the languages that we study. For this
reason, we make the conservative choice to exclude all languages whose
average accuracy was below 75\% from all further analyses below.

\subsection{Irregularity across Languages}\label{sec:irregularity-across}

\begin{figure}
\centering
 \begin{adjustbox}{width=1.0\columnwidth}
\includegraphics{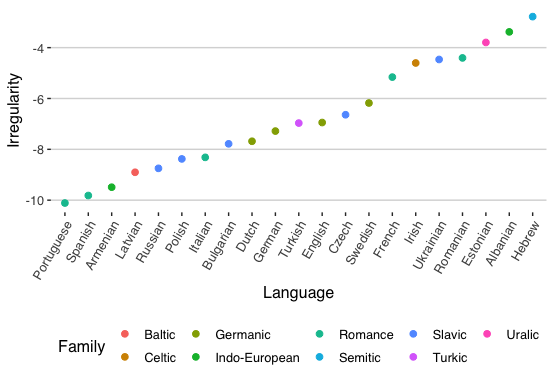}
\end{adjustbox}
\caption{Average degree of irregularity $\iota$ across languages.}
\label{fig:irregularity-per-lang}
\end{figure}

It is often observed  that there are differences in the
prevalence of irregularity across languages
\citep[][]{stolz.t:2012a}. On one end of the spectrum, some languages
have widespread (often suppletive) allomorphy in their marking of
inflectional features. For example, Arabic marks plurality on nouns in
one of more than a dozen different ways and these are idiosyncratic to
the noun stem.  Similarly, Georgian verbs often have different roots
depending on their tense, aspect, or mood marking. On the other end of
the spectrum, it is sometimes claimed that agglutinative languages like
Turkish exhibit no irregularity whatsoever.

\Cref{fig:irregularity-per-lang} displays the average irregularity
score per language for the 21 languages remaining after our 75\%
accuracy criterion. Recall from \cref{eq:log-odds} that the degree of
irregularity $\iota$ is positive when the majority of predicted
probability mass falls on forms that are not the correct target form (i.e.,
the form is irregular), and negative when the majority of probability
mass falls on the predicted form (i.e., the form is regular). As can
be seen from the figure, average irregularity is negative across
languages. This is expected---most forms in these languages are
predicted accurately by the model. However, there is wide variability
in the average irregularity score between languages. In particular,
in the most regular language, Portuguese, correct forms are about
25,000 times more likely on average than alternative forms. In the
most irregular language, Hebrew, correct forms are only about 16 times
more likely on average than alternative forms.  We leave it to future
work to validate and further study these cross-linguistic differences
in irregularity predictions.

\subsection{Irregularity and Frequency}\label{sec:frequency}

In some morphological systems, such as the English past tense, there is a 
strong and well-known correlation between irregularity and frequency  \citep[][]{marcus.g:1992,pinker.s:1999}. In such systems, the most
frequent past forms tend to be irregular and irregular forms tend to
come from the most frequent verbs. Based on cases like this,
it is widely believed in linguistics and psycholinguistics that
there is an association between frequency and irregularity
\citep[][]{bybee.j:1991,haspelmath.m:2010,kiefer.f:2000}. However, to
our knowledge, this relationship  has
never been explicitly tested quantitatively across many languages at once.

Recently, several authors have questioned the received wisdom that
irregularity and frequency are related
\citep[][]{yang.c:2016,fratini.v:2014}.\footnote{But see
  \citet[][]{herce.b:2016}.} Thus, it has become important to test
this relationship empirically.  An example of such a challenge to the
standard assumption comes from \citet[][]{yang.c:2016} who proposed an
influential theory of morphological productivity known as the
\fm{tolerance principle}. The mathematical derivation of the tolerance
principle relies on the assumption that irregular forms are uniformly
distributed throughout the frequency range
\citep[][]{yang.c:2016}.\footnote{Yang tentatively proposes that the
  correlation between frequency and irregularity might be accidental
  in languages such as English. He argues, however, that his theory is
  not contingent on this being the case
  \citep[][pp. 65]{yang.c:2016}.}%

\begin{figure}
\centering
 \begin{adjustbox}{width=1.0\columnwidth}
\includegraphics{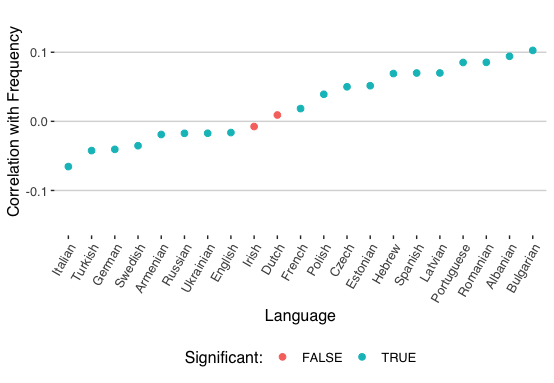}
\end{adjustbox}
\caption{Correlations between irregularity and frequency at the form level.}
\label{fig:frequency-form}
\end{figure}

Here we present the first study to probe the relationship between
irregularity and frequency at scale.  We first examine the
relationship between the degree of irregularity $\iota$ and the
frequency of individual word forms. To study this question, we
examined the Pearson correlation between the log-transformed frequency
of word forms in each language and their predicted irregularity scores
$\iota(w)$. Because word occurrences fall into the class of
\fm{large number of rare event} distributions, finite samples will
tend to underestimate the probability of infrequent words---word forms
that appear $0$ times in some sample often differ by orders of
magnitude in their true probability
\citep[][]{chitashvili.r:1993,baayen.r:2001}. For this reason, we
chose to exclude all frequency $0$ forms from our analyses.

The correlations for the 21 languages considered in this study are
shown in \cref{fig:frequency-form} with significant correlations
($p<0.05$) marked in blue. Overall, a slight trend towards a positive
correlation between irregularity and frequency is discernible in this
set of word forms. Following \citet[][]{mahowald.k:2018}, we tested
this by fitting a mixed-effect model with irregularity as the
dependent variable, language as a random effect (slopes and
intercepts) and log count as a fixed effect
\citep[][]{gelman.a:2007}. The results give a positive coefficient of
$0.064$ for the log count factor. The AIC-corrected log-odds ratio in
favor of the model with a fixed effect of count (compared to a model
with just random effects) is $3.44$. A nested-model likelihood-ratio
$\chi$-squared test shows that the log factor is significant with
$p<0.04$.

\begin{figure}
\centering
 \begin{adjustbox}{width=1.0\columnwidth}
\includegraphics{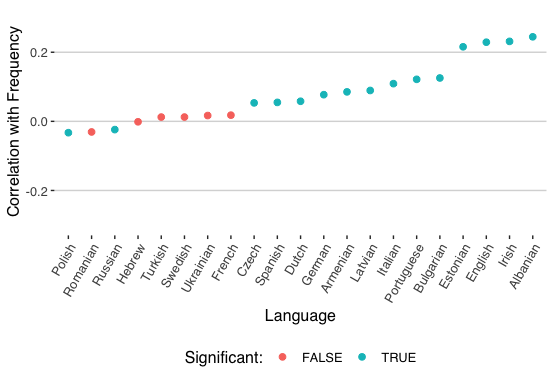}
\end{adjustbox}
\caption{Correlations between irregularity and frequency at the lexeme level.}
\label{fig:frequency-lexeme}
\end{figure}

An important question about irregularity is whether it is a property
of individual forms, or rather whether it inheres to whole paradigms
\citep[][]{baerman.m:2010,stolz.t:2012a,herce.b:2016}. To examine this
question more closely, we ran an alternative correlational analysis
examining the correlation between the sum of the counts of all forms
associated with a lexeme and the average irregularity score for all
forms associated with the lexeme (as in
\cref{eq:avg-irregularity}). \Cref{fig:frequency-lexeme} shows the
results. Overall, a stronger trend towards a positive correlation
between irregularity and frequency is discernible at the lexeme level
than at the word-form level. We tested this by fitting a mixed-effect
model with irregularity as the dependent variable, language as a
random effect (slopes and intercepts) and log count as a fixed effect.
The models gives a positive coefficient of $0.14$ for the log count
factor. The AIC-corrected log-odds ratio in favor of the model with a
fixed effect of count (compared to a model with just random effects)
is $11.8$. A nested-model likelihood-ratio $\chi$-squared test 
shows that the log count factor is significant with $p<0.001$. Thus,
the correlation between irregularity and frequency is considerably
more robust when considered at the lexeme level.

\section{Conclusion}
In this paper, we have introduced a measure of irregularity based on
\word{wug}-testing a model of morphological inflection. In
\cref{sec:validation}, we showed that this measure produces results
that are consistent with human judgements. Focusing on a subset of the
languages for which the model was able to recover the correct
inflected forms at a high rate (\cref{sec:accuracy}), we showed that
average irregularity varies a good deal between languages. This result
is consistent with the findings of \citet[][]{cotterell.r:2018} which
gave large scale empirical evidence of a tradeoff between the size of
morphological paradigms and the predictability of individual forms
within each paradigm.

The main novel empirical result of our paper was presented in
\cref{sec:frequency} which showed that irregularity is correlated with
frequency both at the level of individual forms as well as at the
level of lexemes. To our knowledge, this is the first large-scale
empirical demonstration of this piece of linguistic folk wisdom and
provides evidence relevant to recent proposals questioning this
generalization \citep[][]{fratini.v:2014,yang.c:2016}.

Perhaps of greater interest than this positive result is the
difference in the strength of the correlation between the level of
individual forms and the level of lexemes. This difference appears to
be driven by the fact that, in many cases, lexemes that contain
high-frequency forms will also contain a few low frequency forms as
well. Adopting the terminology of \citet[][]{yang.c:2002}, we can say
that low frequency forms \fm{free-ride} on the higher frequency
members of the lexeme.

This finding lends credence to models of linguistic structure which
group words together by their lexeme or stem. Such models seem
necessary to account for paradigmatic structure cross linguistically
and to deal with phenomena such as the existence of \fm{defective
  paradigms}---the phenomenon whereby certain inflected forms of a
word seem to be impossible for speakers \citep[][]{baerman.m:2010}. A
canonical example is the past participle of \word{stride} (e.g.,
$^*$\word{strode}/$^*$\word{stridden}/$^*$\word{strided}). In these
cases, the problem seems to be that the irregularity of the overall
lexeme is known, but the particular word form has never been
observed. Our results provide further support for the view that
inflected forms represent surface exponence of common underlying
morphological objects.

More generally, we observe that our \word{wug}-test techniques
provides a general way of studying regularity and predictability
within languages and may prove useful for attacking other difficult
problems in the literature, such as detecting inflectional classes. By
measuring which words or lexemes are most predictable from one
another, a general picture of morphological relatedness within a
language can be built in a bottom-up way. 

\section*{Acknowledgments}
The third author gratefully acknowledges support from the Fonds de Recherche du Qu\'{e}bec---Soci\'{e}t\'{e} et Culture and the Natural Sciences and Engineering Research Council of Canada.

\bibliography{main.bbl}

\begin{thebibliography}{39}
\expandafter\ifx\csname natexlab\endcsname\relax\def\natexlab#1{#1}\fi

\bibitem[{Ackerman and Malouf(2013)}]{ackerman.f:2013}
Farrell Ackerman and Robert Malouf. 2013.
\newblock Morphological organization: {T}he low conditional entropy conjecture.
\newblock \emph{Language}, 89(3):429--464.

\bibitem[{Albright and Hayes(2003)}]{albright2003rules}
Adam Albright and Bruce Hayes. 2003.
\newblock Rules vs. analogy in {E}nglish past tenses: {A}
  computational/experimental study.
\newblock \emph{Cognition}, 90(2):119--161.

\bibitem[{Baayen(2001)}]{baayen.r:2001}
R.~Harald Baayen. 2001.
\newblock \emph{Word Frequency Distributions}.
\newblock Springer, Berlin, Germany.

\bibitem[{Baerman et~al.(2015)Baerman, Brown, and Corbett}]{baerman.m:2015}
Matthew Baerman, Dunstan Brown, and Greville~G. Corbett. 2015.
\newblock \emph{Understanding and measuring morphological complexity: An
  introduction.}
\newblock Oxford University Press.

\bibitem[{Baerman et~al.(2010)Baerman, Corbett, and Brown}]{baerman.m:2010}
Matthew Baerman, Greville~G. Corbett, and D.~P. Brown. 2010.
\newblock \emph{Defective Paradigms: Missing forms and what they tell us}.
\newblock Oxford University Press, Oxford, England.

\bibitem[{Berko(1958)}]{berko.j:1958}
Jean Berko. 1958.
\newblock The child's learning of {E}nglish morphology.
\newblock \emph{Word}, 14:150--177.

\bibitem[{Bybee(1985)}]{bybee.j:1985}
Joan~L. Bybee. 1985.
\newblock \emph{Morphology: {A} Study of the Relation between Meaning and
  Form}.
\newblock John Benjamins, Amsterdam.

\bibitem[{Bybee(1991)}]{bybee.j:1991}
Joan~L. Bybee. 1991.
\newblock Natural morphology: {T}he organization of paradigms and language
  acquisition.
\newblock In Thom Huebner and Charles~A. Ferguson, editors, \emph{Cross
  Currents in Second Language Acquisition and Linguistic Theory}. John
  Benjamins Publishing Company.

\bibitem[{Chitashvili and Baayen(1993)}]{chitashvili.r:1993}
Revas~J. Chitashvili and R.~Harald Baayen. 1993.
\newblock Word frequency distributions.
\newblock \emph{Quantitative Text Analysis}, pages 54--135.

\bibitem[{Cotterell et~al.(2018{\natexlab{a}})Cotterell, Kirov, Hulden, and
  Eisner}]{cotterell.r:2018}
Ryan Cotterell, Christo Kirov, Mans Hulden, and Jason Eisner.
  2018{\natexlab{a}}.
\newblock On the complexity and typology of inflectional morphological systems.
\newblock \emph{Transaction of the Association for Computational Linguistics
  ({TACL})}.

\bibitem[{Cotterell et~al.(2018{\natexlab{b}})Cotterell, Kirov, Mielke, and
  Eisner}]{N18-2087}
Ryan Cotterell, Christo Kirov, Sebastian~J. Mielke, and Jason Eisner.
  2018{\natexlab{b}}.
\newblock \href {https://doi.org/10.18653/v1/N18-2087} {Unsupervised
  disambiguation of syncretism in inflected lexicons}.
\newblock In \emph{Proceedings of the 2018 Conference of the North American
  Chapter of the Association for Computational Linguistics: Human Language
  Technologies, Volume 2 (Short Papers)}, pages 548--553. Association for
  Computational Linguistics.

\bibitem[{Cotterell et~al.(2017{\natexlab{a}})Cotterell, Kirov, Sylak-Glassman,
  Walther, Vylomova, Xia, Faruqui, K{\"u}bler, Yarowsky, Eisner, and
  Hulden}]{K17-2001}
Ryan Cotterell, Christo Kirov, John Sylak-Glassman, G{\.{e}}raldine Walther,
  Ekaterina Vylomova, Patrick Xia, Manaal Faruqui, Sandra K{\"u}bler, David
  Yarowsky, Jason Eisner, and Mans Hulden. 2017{\natexlab{a}}.
\newblock \href {https://doi.org/10.18653/v1/K17-2001} {{CoNLL-SIGMORPHON} 2017
  shared task: {U}niversal morphological reinflection in 52 languages}.
\newblock In \emph{Proceedings of the CoNLL SIGMORPHON 2017 Shared Task:
  Universal Morphological Reinflection}, pages 1--30. Association for
  Computational Linguistics.

\bibitem[{Cotterell et~al.(2017{\natexlab{b}})Cotterell, Sylak-Glassman, and
  Kirov}]{cotterell.r:2017}
Ryan Cotterell, John Sylak-Glassman, and Christo Kirov. 2017{\natexlab{b}}.
\newblock Neural graphical models over strings for principal parts
  morphological paradigm completion.
\newblock In \emph{Proceedings of the 15th Conference of the {E}uropean Chapter
  of the Association for Computational Linguistics ({EACL2017})}.

\bibitem[{Fratini et~al.(2014)Fratini, Acha, and Laka}]{fratini.v:2014}
Viviana Fratini, Joana Acha, and Itziar Laka. 2014.
\newblock Frequency and morphological irregularity are independent variables.
  {E}vidence from a corpus study of {S}panish verbs.
\newblock \emph{Corpus Linguistics and Linguistic Theory}, 10(2):289 --314.

\bibitem[{Gelman and Hill(2007)}]{gelman.a:2007}
Andrew Gelman and Jennifer Hill. 2007.
\newblock \emph{Data Analysis using Regression and Multilevel/Hierarchical
  Models}.
\newblock Cambridge University Press, Cambridge.

\bibitem[{Haspelmath and Sims(2010)}]{haspelmath.m:2010}
Martin Haspelmath and Andrea~D. Sims. 2010.
\newblock \emph{Understanding Morphology}.
\newblock Hodder Education.

\bibitem[{Hay(2003)}]{hay.j:2003}
Jennifer Hay. 2003.
\newblock \emph{Causes and Consequences of Word Structure}.
\newblock Routledge, New York, NY.

\bibitem[{Herce(2016)}]{herce.b:2016}
Borja Herce. 2016.
\newblock Why frequency and morphological irregularity are not independent
  variables in {S}panish: {A} response to {F}ratini et al. (2014).
\newblock \emph{Corpus Linguistics and Linguistic Theory}, 12(2).

\bibitem[{Hockett(1954)}]{hockett.c:1954}
Charles~F. Hockett. 1954.
\newblock Two models of grammatical description.
\newblock \emph{Word}, 10:210--231.

\bibitem[{Kiefer(2000)}]{kiefer.f:2000}
Ferenc Kiefer. 2000.
\newblock Regularity.
\newblock In \emph{Morphologie: Ein internationales Handbuch zur Flexion und
  Wortbildung/Morphology: An international Handbook on Inflection and
  Word-Formation}. Walter {d}e Gruyter, Berlin.

\bibitem[{Kirov et~al.(2018)Kirov, Cotterell, Sylak-Glassman, Walther,
  Vylomova, Xia, Faruqui, Mielke, McCarthy, K{\"u}bler
  et~al.}]{kirov2018unimorph}
Christo Kirov, Ryan Cotterell, John Sylak-Glassman, G{\'e}raldine Walther,
  Ekaterina Vylomova, Patrick Xia, Manaal Faruqui, Sebastian Mielke, Arya~D
  McCarthy, Sandra K{\"u}bler, et~al. 2018.
\newblock Unimorph 2.0: {U}niversal morphology.
\newblock \emph{arXiv preprint arXiv:1810.11101}.

\bibitem[{Mahowald et~al.(2018)Mahowald, Dautriche, Gibson, and
  Piantadosi}]{mahowald.k:2018}
Kyle Mahowald, Isabelle Dautriche, Edward Gibson, and Steven~Thomas Piantadosi.
  2018.
\newblock Word forms are structured for efficient use.
\newblock \emph{Cognitive Science}, 42(8):3116--3134.

\bibitem[{Marcus et~al.(1992)Marcus, Pinker, Ullman, Hollander, Rosen, and
  Xu}]{marcus.g:1992}
Gary~F. Marcus, Steven Pinker, Michael~T. Ullman, Michelle Hollander, T.~John
  Rosen, and Fei Xu. 1992.
\newblock \emph{Overregularization in Language Acquisition}.
\newblock Monographs of the society for research in child development.
  University of Chicago Press, Chicago, IL.

\bibitem[{McClelland and Patterson(2002{\natexlab{a}})}]{mcclelland.j:2002}
James~L. McClelland and Karalyn Patterson. 2002{\natexlab{a}}.
\newblock Rules or connections in past-tense inflections: {W}hat does the
  evidence rule out?
\newblock \emph{Trends in Cognitive Sciences}, 6(11):465--472.

\bibitem[{McClelland and Patterson(2002{\natexlab{b}})}]{mcclelland.j:2002a}
James~L. McClelland and Karalyn Patterson. 2002{\natexlab{b}}.
\newblock `{W}ords or {R}ules' cannot exploit the regularity in exceptions.
\newblock \emph{Trends in Cognitive Sciences}, 6(11):464--465.

\bibitem[{O'Donnell(2015)}]{odonnell.t:2015}
Timothy~J. O'Donnell. 2015.
\newblock \emph{Productivity and Reuse in Language: {A} Theory of Linguistic
  Computation and Storage}.
\newblock The MIT Press, Cambridge, Massachusetts.

\bibitem[{Pinker(1999)}]{pinker.s:1999}
Steven Pinker. 1999.
\newblock \emph{Words and Rules}.
\newblock HarperCollins, New York, NY.

\bibitem[{Pinker and Prince(1988)}]{pinker.s:1988}
Steven Pinker and Alan Prince. 1988.
\newblock On language and connectionism: {A}nalysis of a parallel distributed
  processing model of language acquisition.
\newblock \emph{Cognition}, 28:73--193.

\bibitem[{Pinker and Ullman(2002{\natexlab{a}})}]{pinker.s:2002a}
Steven Pinker and Michael~T. Ullman. 2002{\natexlab{a}}.
\newblock Combination and structure, not gradedness, is the issue.
\newblock \emph{Trends in Cognitive Sciences}, 6(11):472--474.

\bibitem[{Pinker and Ullman(2002{\natexlab{b}})}]{pinker.s:2002}
Steven Pinker and Michael~T. Ullman. 2002{\natexlab{b}}.
\newblock The past and future of the past tense debate.
\newblock \emph{Trends in Cognitive Sciences}, 6(11):456--463.

\bibitem[{Prasada and Pinker(1993)}]{prasada.s:1993}
Sandeep Prasada and Steven Pinker. 1993.
\newblock Generalisation of regular and irregular morphological patterns.
\newblock \emph{Language and Cognitive Processes}, 8(1):1--56.

\bibitem[{Rabiner(1989)}]{rabiner1989tutorial}
Lawrence~R. Rabiner. 1989.
\newblock A tutorial on hidden {M}arkov models and selected applications in
  speech recognition.
\newblock \emph{Proceedings of the IEEE}, 77(2):257--286.

\bibitem[{Rumelhart and McClelland(1986)}]{rumelhart.d:1986}
David~E. Rumelhart and James~L. McClelland. 1986.
\newblock On learning the past tenses of {E}nglish verbs.
\newblock In \emph{Parallel Distributed Processing: Explorations in the
  Microstructure of Cognition.}, volume~2, pages 216--271. Bradford Books/MIT
  Press, Cambridge, MA.

\bibitem[{Siegel(1974)}]{siegel.d:1974}
Dorothy Siegel. 1974.
\newblock \emph{Topics in English Morphology}.
\newblock Ph.D. thesis, Massachusetts Institute of Technology.

\bibitem[{Stolz et~al.(2012)Stolz, Otsuka, Urdze, and {v}an~{d}er
  Auwera}]{stolz.t:2012a}
Thomas Stolz, Hitomi Otsuka, Aina Urdze, and Johan {v}an~{d}er Auwera. 2012.
\newblock Introduction: {I}rregularity --- glimpses of a ubiquitous phenomenon.
\newblock In Thomas Stolz, Hitomi Otsuka, Aina Urdze, and Johan {v}an~{d}er
  Auwera, editors, \emph{Irregularity in Morphology (and Beyond)}, pages 7--38.
  Akademie Verlag, Berlin, Germany.

\bibitem[{Stump(2001)}]{stump.g:2001}
Gregory~T. Stump. 2001.
\newblock Inflection.
\newblock In \emph{Handbook of Morphology}. Blackwell, Oxford, England.

\bibitem[{Wu and Cotterell(2019)}]{wu2019exact}
Shijie Wu and Ryan Cotterell. 2019.
\newblock Exact hard monotonic attention for character-level transduction.
\newblock \emph{arXiv preprint arXiv:1905.06319}.

\bibitem[{Yang(2002)}]{yang.c:2002}
Charles~D. Yang. 2002.
\newblock \emph{Knowledge and Learning in Natural Language}.
\newblock Oxford linguistics. Oxford University Press, New York.

\bibitem[{Yang(2016)}]{yang.c:2016}
Charles~D. Yang. 2016.
\newblock \emph{The Price of Productivity: {H}ow Children Learn to Break the
  Rules of Language}.
\newblock The MIT Press, Cambridge, Massachusetts.

\end{thebibliography}
\bibliographystyle{acl_natbib}

\end{document}